\RequirePackage{fix-cm}
\documentclass{svjour3}                     
\smartqed  
\usepackage[top=1.1in, bottom=1.2in, left=1.2in, right=1.2in]{geometry} 
\usepackage{graphicx}
\usepackage[numbers]{natbib} 
\journalname{Neural Computing and Applications}

\usepackage{multirow}
\usepackage{makecell}
\usepackage[usedvipsnames]{xcolor}
\usepackage{siunitx}
\usepackage{times}
\usepackage{textcomp}
\usepackage{hyperref}
\usepackage{caption}

\newcommand{\eg}{\textit{e.g.} }
\newcommand{\ie}{\textit{i.e.} }

\newcommand{\rev}[1]{\textcolor{black}{#1}}

\usepackage[numbers]{natbib}

\begin{document}

\title{\rev{RealHePoNet: a robust single-stage ConvNet for head pose estimation in the wild}}



\author{Rafael Berral-Soler         \and
        Francisco J. Madrid-Cuevas   \and
        Rafael Mu\~noz-Salinas
        \and
        Manuel J. Mar\'in-Jim\'enez
}


\institute{Rafael Berral-Soler, 
Francisco J. Madrid-Cuevas,
Rafael Mu\~noz-Salinas, 
Manuel J. Mar\'in-Jim\'enez
\at
Department of Computing and Numerical Analysis, University of Cordoba, Spain \\
            \email{(corresponding author) mjmarin AT uco.es}
}

\date{Received: date / Accepted: date}

\maketitle

\begin{abstract}

Human head pose estimation in images has applications in many fields such as human-computer interaction or video surveillance tasks. In this work, we address this problem, defined here as the estimation of both vertical (tilt/pitch) and horizontal (pan/yaw) angles, through the use of a single \rev{Convolutional Neural Network (ConvNet) model}, trying to balance precision and inference speed in order to maximize its usability in real-world applications.
Our model is trained over the combination of two datasets: `Pointing'04' (aiming at covering a wide range of poses) and `Annotated Facial Landmarks in the Wild' (in order to improve robustness of our model for its use on real-world images). Three different partitions of the combined dataset are defined and used for training, validation and testing purposes.
As a result of this work, we have obtained a trained ConvNet model, coined RealHePoNet, \rev{that given a low-resolution grayscale input image, and without the need of using facial landmarks,} is able to estimate with low error both tilt and pan angles (\texttildelow4.4\si{\degree} average error on the test partition). Also, given its low inference time (\texttildelow6 ms per head), we consider our model usable even when paired with medium-spec hardware (\ie GTX 1060 GPU).\\
\noindent Code available at: \url{https://github.com/rafabs97/headpose_final} \\
\noindent Demo video at: \url{https://www.youtube.com/watch?v=2UeuXh5DjAE}

\end{abstract}


\section{Introduction} \label{s:intro}
\begin{figure}[t]
	\centering
	\includegraphics[width=1.0 \textwidth]{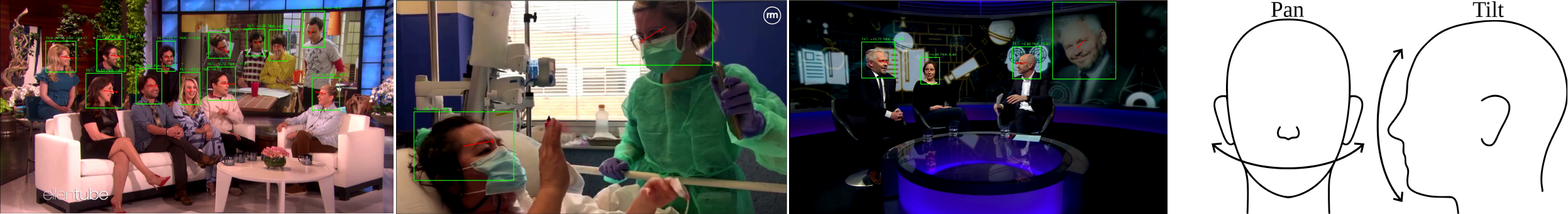}
	\caption{\textbf{Objective of this paper}. The goal of this paper is to develop a head pose estimator that works on unconstrained images. \textbf{(left)} Actual outputs of our estimator on challenging test videos from YouTube (\url{https://www.youtube.com/watch?v=2UeuXh5DjAE}). Note that neither the head detector, nor the pose estimator were trained on images with face-masks. \textbf{(right)} Head pose is mainly defined by pan and tilt angles. (Best viewed in digital format.) }
	\label{fig:teaser}
\end{figure}

Given a human head detected in a picture, we can define the task of \textit{head pose estimation} (HPE) as the estimation, relative to the camera, of both vertical (tilt/pitch) and horizontal (pan/yaw) angles (see Figure \ref{fig:teaser}) -- a third angle (roll) can be estimated, but it falls outside the scope of this work.

Human head pose estimation is useful in many situations: for instance in vehicles (detecting if the driver of a vehicle is paying attention to the road \cite{Chutorian2010}), human-computer interaction (detecting where the user's attention is being drawn \cite{Vatahska2007}), social interaction understanding (detecting if people is looking at each other~\cite{Marin2019}), video surveillance systems \cite{Gourier2006, Pereira2017} or to aid various aerial cinematography tasks \cite{Passalis2020}.

In previous years, multiple techniques \cite{Chutorian2009, czupry2014survey} have been proposed in order to deal with HPE. Usually these techniques involve the use of special hardware (e.g. depth cameras \cite{Fanelli2011B}) \rev{, what limits their applicability in real scenarios (\eg regular smartphones), or requires the (prone-error) estimation of facial landmarks followed by a 3D head model fitting~\cite{zhu2017pami}, what greatly increases the amount of computational resources needed to perform the estimation.} 
\rev{Nowadays, it is well-known that the use of deep learning (DL) has been adopted in many fields, due to its success in different real-world applications~\cite{wijnands2019ncaa,zhang2020joint}. Among all existing DL-based approaches, when the input is a digital signal (\eg sound tracks or images), the Convolutional Neural Networks (ConvNets) have shown to be very effective, if the amount of available training data is big enough.}
A ConvNet is an extension of the concept of Artificial Neural Network including, among others, one or more convolutional layers intended to extract local features from raw input data; the neurons in these layers act as a convolutional filter over the output of the previous layer. ConvNets~\cite{Lecun1998} have been used to tackle many problems related to Computer Vision such as 
character recognition \cite{Yuan2012}, image categorization~\cite{simonyan2015iclr}, people identification~\cite{castro2020ncaa}, object detection~\cite{Liu2015} or human pose estimation~\cite{marin2018jvci}, among others.
%


Multiple efforts have been made on solving the head pose estimation problem using ConvNets~\cite{Liu2016,Patacchiola2017,Passalis2020}. However these efforts have usually implied the use of fully synthetic datasets~\cite{Liu2016}; others require the use of different models in order to estimate different angles \cite{Patacchiola2017} \rev{or the use of different specialized network branches \cite{ruiz2018hopenet}}; and, \rev{some others need the fitting of a 3D head model~\cite{zhu2017pami} to obtain the estimation.}
In this work we try to solve the problem of head pose estimation over low-resolution, grayscale, real world images using a single ConvNet model, aiming to get a good precision (\ie low error) while keeping the estimation time as low as possible. 
\rev{Therefore, in order to overcome the limitations of the existing methods, we propose in this work a single ConvNet (in contrast to approaches with a model per angle~\cite{Patacchiola2017} or branch per angle~\cite{ruiz2018hopenet}), working on grayscale low resolution images (\ie $64\times64$ vs $224\times224 \times 3$~\cite{ruiz2018hopenet}), not requiring the detection of facial landmarks~\cite{zhu2017pami}; and, with a modest amount of parameters (6.8M vs 23.9M~\cite{ruiz2018hopenet}), allowing a possible future deployment on mobile devices.}

\begin{figure*}[b]
	\centering
	\includegraphics[width=\textwidth]{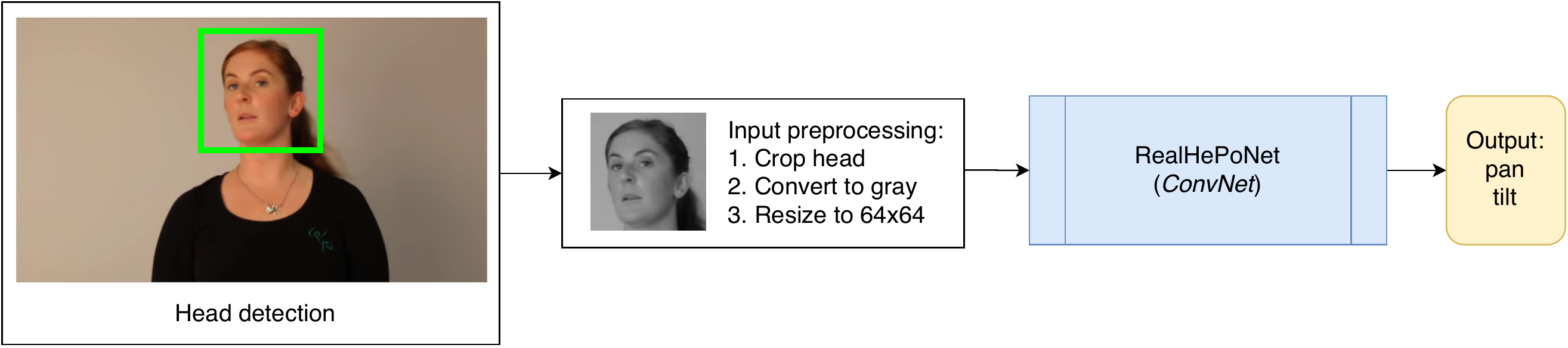}
	\caption{\rev{\textbf{Flowchart of the proposed model.} Given an input image, the first step is to detect all the visible human heads. Then, each head is independently processed: (i) crop the target head, (ii) convert to grayscale and (iii) resize to $64\times 64$ pixels. The resulting image is used as the input of our ConvNet. The output is the estimation of both the pan and tilt angles. (Best viewed in digital format).}}
	\label{fig:flowchart}
\end{figure*}

%
Therefore, the main contribution of this paper can be summarized as the development of a human head pose estimator using a single neural network model, balancing both a reasonably low error with a low inference time, and achieved through the use of a simple, iterative, architecture and data augmentation parameter search method.
\rev{The main steps of our approach for head pose estimation on images are summarized in Figure~\ref{fig:flowchart}. Note that no computationally expensive preprocessing steps (\eg facial landmark detection) are needed.}
As a result, the software developed during the course of this work and the final trained models are publicly available in a GitHub repository \cite{headpose_final}.

The rest of this paper is organized as follows. In Section \ref{s:relwork} we summarize some of the previous attempts done in order to solve HPE. Section \ref{s:dataset_processing} describes the process used in this work in order to obtain useful data for training our HPE models. %
In Section \ref{s:architectures} we explain the different architectures we have tested in order to solve the problem. In Section \ref{s:experiments} we present the experiments performed to obtain a competitive model, whose results are discussed in Section \ref{s:results}. Finally, in Section \ref{s:conclusion} we extract some conclusions and suggest future lines of research.

\section{Related work} \label{s:relwork}
The problem we address in this paper is head pose estimation. In this section we summarize some of the previous works done in order to solve this problem. We divide them into two groups: classic techniques (not using Deep Learning) and Deep Learning-based techniques.

\subsection{Classic techniques}
The following techniques do not use Deep Learning in order to estimate head pose angles. Raytchev et al. in 2004 \cite{Raytchev2004} treated the problem as a dimensionality reduction problem, using Isomap \cite{Tenenbaum2000} in order to get the three relevant components (pan, tilt and roll angles) out of a greater dimensionality input (the given picture). Ba et al., also in 2004 \cite{Ba2004}, proposed a technique capable of jointly tracking head position and head pose estimation using particle filters, a type of Monte Carlo algorithm. Then, Chutorian et al. in 2007 \cite{Chutorian2007} presented a method that used Support Vector Machines (SVMs) over a localized gradient histogram of input images in order to obtain functions capable of estimating every pose angle (pan, tilt and roll). That same approach was used as a verification method in 2010 \cite{Chutorian2010}, where a method similar to the one in the paper by Ba et al. is used to track pose values after being given their initial values, also obtained with this method. Fanelli et al. in 2011 \cite{Fanelli2011} estimated head pose using random forests applied to a regression problem, having the disadvantage of requiring special hardware (depth cameras); however, in a later work, this technique was used over the data provided by a Kinect sensor \cite{Fanelli2011B}.
Then, Marín-Jiménez et al.~\cite{Marin2014} trained two Gaussian Process regressors (pan and tilt angles) for head pose estimation using as input the HOG (Histogram of Oriented Gradients) descriptors of head images.
On the other hand, in a more restrictive setup, Mu\~noz-Salinas et al. \cite{Salinas2012} propose a multiview approach for head pose estimation where individual classifiers are run on different views of the same person, and the information is fused considering that the three-dimensional position of each camera in the environment is known.

We also consider as classic techniques those using neural networks as part of the estimation process, as long as they are not deep neural networks. Gourier et al. in 2006 \cite{Gourier2006} used multiple linear auto-associative memories (a type of neural network) trained over different poses in order to estimate the pose from a new input. Balasubramanian et al. in 2007 \cite{Balasubramanian2007} used an approach similar to the work by Raytchev et al. \cite{Raytchev2004}, but using empirical pose values in order to bias the non-linear fitting; the mapping from the original dimensionality space to the reduced dimensionality space is done using Radial Basis Functions (RBF). Vatashka et al., also in 2007 \cite{Vatahska2007}, presented a method of pose estimation from monocular pictures, using a neural network to estimate the pose from the detected position of multiple facial landmarks obtained using AdaBoost classifiers, where two neural networks were used: one for frontal faces and other for left/right profiles.

\rev{Recently, several works using facial landmarks have been proposed. Abate et al.~\cite{Abate2019}, in 2019, proposed to use a quadtree encoding of the detected landmarks to obtain a feature vector that is classified on a set of prototype poses. This method was improved later in 2020 \cite{Barra2020} by using a web-shaped encoding. Yuan et al.~\cite{Yuan2020pr}, in 2020,  proposed to obtain the head pose by calculating the adjustment of the detected feature points to a predefined model.
In contrast to these models, ours does not require the computation of facial landmarks, eliminating the need of this pre-processing step.}

\subsection{Deep learning techniques}

Methods based on classic techniques usually require obtaining image ``descriptors'' before making a prediction. These descriptors are metrics obtained from the image and provide us with information about color, texture, gradient, etc. Deep Learning methods usually are able to directly extract image descriptors on their own, while obtaining better results than classic techniques \cite{Lathuiliere2018}; however, they usually require far larger datasets in order to obtain good results \cite{Patacchiola2017}.

Liu et al. in 2016 \cite{Liu2016} used a method based only on a single Convolutional Neural Network composed by three convolutional blocks followed by three fully connected layers. Using as input the picture of the head and returning as output the values for all three angles of the pose (pan, tilt and roll); they trained their models only on synthetic datasets. This could be problematic when testing the model on real-world conditions. The opposite has not been proved, as they only tested the model over datasets obtained in laboratory conditions. In contrast, our goal is to develop a system able to work in unconstrained real-world conditions.

An interesting feature of neural networks is that part of a previously trained model can be adapted and modified to be used in another similar problem. Lathuiliere et al. in 2017 \cite{Lathuiliere2017} proposed a method that, based on a previous architecture, adds a new output layer implementing a Gaussian mixture of inverse linear regressions in order to estimate pose values, combined with a ``fine tuning'' of the network to reduce the number of parameters to adjust. The VGG-16 architecture \cite{simonyan2015iclr} was used as backbone; while the obtained precision is good, such a large model could severely hinder inference time on common use hardware, making it not suitable for real-time applications. Note that our final model (RealHePoNet) has roughly 6.8 million parameters, in contrast to the 138 million parameters of VGG-16. 

Neural network models are heavily influenced by hyperparameter values during training; as such, adaptive gradient methods can be very useful. Patacchiola et al. in 2017 \cite{Patacchiola2017} compared the performance of multiple architectures in combination with multiple adaptive gradient methods over the head pose estimation problem. Unlike Liu et al. \cite{Liu2016}, the models were trained on a real, non-synthetic dataset, containing a greater diversity of age, ethnicity, expressions, occlusions, etc. and thus being more representative of real-world conditions. However, they used three different models in order to predict each angle (tilt, pan, roll), potentially requiring more time than needed in order to estimate the pose, and they used a face detector instead of a head detector (\ie our case), making it impossible for the system to obtain an estimation on partially occluded faces.
\rev{Then, in 2018, Ruiz et al.~\cite{ruiz2018hopenet} presented a two-stage ConvNet model able to estimate the head pose directly from RGB images. Their model starts by classifying the pose in a rough manner and, then, the pose is further refined by regression. In contrast to this approach, we propose a single-stage model that works on single-channel (\ie grayscale) images.}

\rev{Recently, in 2019, Xia et al.~\cite{Xia2019} proposed a hybrid method which uses a heat map, obtained from a predefined set of detected landmarks, to guide a ConvNet to focus on the areas of the image where to extract the facial descriptors.}
\rev{On the other hand, Zhu et al.~\cite{zhu2017pami} introduced a framework that combines facial landmarks, ConvNets and a 3D morphable model to estimate the head pose. 
However, our proposed model can directly estimate the head pose without requiring any head model fitting or facial landmark computation.}


\section{Dataset pre-processing} \label{s:dataset_processing}

\rev{In this section, after describing the datasets used for training and evaluating the proposed model, we explain the pre-processing steps applied to the original datasets in order to get useful data for model training.}

\subsection{\rev{Head pose datasets}} \label{subsec:datasets}
%
\begin{figure}[t]
	\centering
	\includegraphics[width=0.5 \textwidth]{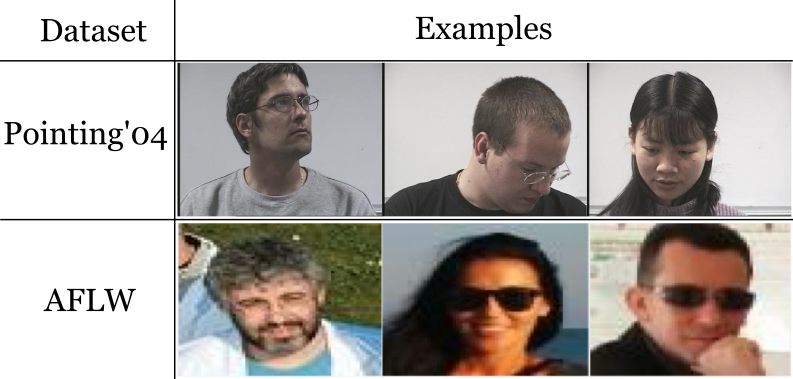}
    \caption{\textbf{Example pictures from annotated head pose datasets.} AFLW contains images taken in the wild, whereas Pointing'04 is composed of images taken under controlled conditions.}
	\label{fig:example_pictures}
\end{figure}

\rev{We use two public datasets for training our models (see Figure \ref{fig:example_pictures}): Pointing'04 \cite{Pointing04} and `Annotated Facial Landmarks in the Wild' (AFLW)~\cite{Koestinger2011}. 
Note that, as pictures from the AFLW dataset cannot be published for licensing reasons, the pictures depicted in the bottom row of Figure \ref{fig:example_pictures} are just some examples similar to the ones appearing in the AFLW dataset, obtained from the paper by Patacchiola et al. \cite{Patacchiola2017}}.

\noindent\textbf{Pointing'04 } This dataset is composed by mugshot-like pictures of 15 individuals: 13 male and 2 female. For each individual we have 2 series, each one containing 93 pictures. The poses in the dataset result from the combination of one in a list of discrete values for the tilt angle (-90º, -60º, -30º, -15º, 0º, 15º, 30º, 60º and 90º) and other value for the pan angle (-90º, -75º, -60º, -45º, -30º, -15º, 0º, 15º, 30º, 45º, 60º, 75º and 90º). Each picture contains only the head of the photographed person, the background is neutral, there is little change in facial accessories, and there are no variations in lighting conditions. As such, while this dataset is useful in terms of the amount of poses appearing on it, training only on this dataset may lead to a biased model, having problems to predict the pose values for a picture obtained in different conditions.\\
    
\noindent\textbf{AFLW } This dataset contains a large number of pictures uploaded by Flickr \cite{Flickr} users. The entire dataset contains about $25,000$ pictures; however, for each picture there can be more than one individual, making the number of heads appearing on the dataset larger than this value. 
We have used only a fraction of the dataset containing $21,123$ pictures. By gender, 41\% of the individuals are male and 59\% female. The pose distribution in this dataset is non-uniform, as there are pose ranges appearing way more than others, following user preferences; poses range between -45º and 45º for the tilt angle, and between -100º to +100º for the pan angle. Since this dataset contains pictures uploaded by users, there can be multiple persons in each picture, and the amount of variation in conditions such as age, race, expressions or clothing is much larger in comparison with the Pointing'04 dataset; as such, training over this dataset allows to get more robust models. 
\rev{For each annotated person, the corresponding coarse head pose is provided by the authors of the dataset, obtained by fitting a mean 3D face with the POSIT algorithm.}
    
\begin{figure}[t]
	\centering
	\includegraphics[scale=0.3]{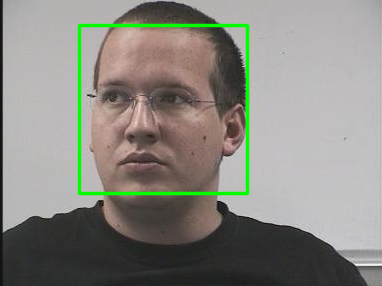}
	\includegraphics[scale=0.3]{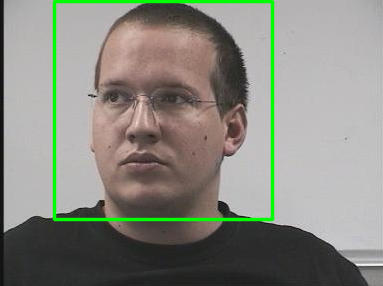}
	\caption{Cropping types: From shorter side (left), and from longer side (right).}
	\label{fig:cropping}
\end{figure}

\subsection{Obtaining human heads from images} \label{ss:basic_operations}
As our aim is the development of a model able to work in real and challenging image scenarios (as shown in Figure~\ref{fig:teaser}), in contrast to other existing works that detect only faces~\cite{Patacchiola2017, Gourier2006}, we detect full human heads, what allows a higher recall in terms of people detection in complex scenes~\cite{Marin2019}. Therefore, we provide below details on this process.

\subsubsection{Head detection} \label{sss:head_detection}
In order to get data suitable to be used as input for our model, we need a way to detect every human head appearing in a picture. In this work we use for this task the head detector model trained in the work by Mar\'in-Jim\'enez et al. \cite{Marin2019}, which allows us to precisely detect heads using neural networks. This model is based on the \textit{Single Shot Detector} (SSD) architecture \cite{Liu2015} and, for each input picture, outputs a list of bounding boxes, each one containing a head.

The procedure for detecting heads from an input picture is the following:
\begin{enumerate}
    \item Resize the input picture to the SSD model's input size ($512 \times 512$ pixels).
    \item Obtain the list $\mathcal{L}$ of predicted bounding boxes and their corresponding confidence score (normalized to $[0,1]$) from the input image.
    \item Filter the list of bounding boxes $\mathcal{L}$ using the confidence value as criterion; we discard every bounding box with a confidence value lower than a certain ``confidence threshold'' ($CT$), selected by cross-validation.
    \item Transform back the coordinates of the selected bounding boxes to their coordinates in the original (before resizing) picture.
\end{enumerate}

\subsubsection{Head cropping} \label{sss:cropped_picture_extraction}
Once we have obtained every valid detection for a given input image, we can use them to extract image crops from the picture, each one corresponding to the region defined by the bounding box. These crops are the ones used as input for our head pose estimator.

The process to obtain valid crops from the bounding boxes obtained in the previous step are the following:
\begin{enumerate}
    \item Note that in order to reduce distortion in the input picture for the model, bounding boxes should be square shaped. For this reason we reduce the original bounding boxes, keeping only the square section with a side length equal to the length of the shortest side of the original bounding box. Using the shortest side increases the probability of the crop of being completely contained inside the original picture, while maintaining most of the head contained inside the box, thus making possible to estimate the orientation (see Figure \ref{fig:cropping} for comparison).
    \item If the predicted bounding box is outside the limits of the input picture, we try to shift it just enough to fit it inside the picture limits. After this step, if the bounding box is still outside the image limits, we discard it.
    \item Finally, the remaining bounding boxes are used to obtain cropped pictures from the original picture; these crops can be used as input for the pose estimator, either for training or for estimating the pose for the head in the picture. Those crops will be stored as grayscale pictures in order to be used as the input for our models.
\end{enumerate}

\subsection{Preparation of the datasets} \label{ss:processing_strategy}

\subsubsection{AFLW dataset} \label{sss:aflw_dataset}

The first dataset we processed is the AFLW dataset, the largest dataset used in this work.
The process for obtaining crops from the AFLW dataset is the following:

For each image file in the dataset:
\begin{enumerate}
    \item We obtain a list of bounding boxes by using the previously mentioned head detector (Sec.~\ref{sss:head_detection}).
    \item After that, we try to find a match between those predicted bounding boxes and a ground-truth bounding box annotation, using Intersection-over-Union \cite{Rosebrock2016} (IoU) as the metric for the correspondence between bounding boxes: each ground-truth bounding box will be matched with the detected bounding box that obtains the highest IoU value when overlapped with it. 
    Unmatched bounding boxes will be discarded. 
    \item Detected bounding boxes with a match will be used to obtain crops from the original picture. These crops and its horizontally flipped version will be stored, along with their ground-truth tilt and pan values. The pose value for each detected bounding box will be the one assigned to the actual ground-truth bounding box matching with it.
\end{enumerate}

\subsubsection{Pointing'04 dataset} \label{sss:pointing04_dataset}
We explain below the process used for obtaining crops from this dataset.
For each image file in the dataset:
\begin{enumerate}
    \item We obtain a list of bounding boxes using the detector model; ideally, there should be only one bounding box per picture.
    \item From the tilt and pan values for the picture we can obtain an ``angle-class'' label (given that the angle values in Pointing'04 dataset are discrete). 
    \item Using the detected bounding boxes, we can obtain crops from the original picture. We store the cropped image window and its corresponding pose values alongside the mirrored version of the obtained crop (as we do for the AFLW dataset). We keep a copy of these two images and their respective class values in memory.
\end{enumerate}

During training, we observed that balancing both the number of samples used from each head pose (discretized) and each dataset is beneficial in terms of model generalization. Therefore, we use this strategy during model training.


\subsection{Confidence threshold selection} \label{ss:confidence_threshold}
The head detector used in this work outputs its detections alongside a confidence value for each detection (\ie detection score). This value indicates the level of confidence the detector has on a certain detection containing the desired object (in this case, a human head), being 0 the minimum value (no confidence at all) and 1 the maximum value (full confidence). In order to discern false detections from true detections we use a parameter called confidence threshold ($CT$). This parameter determines how restrictive the filter will be when considering a detection as valid: we discard any detection with a confidence value lower than this threshold. We have selected the value for this parameter based on three metrics:
\begin{itemize}
    \item Number of valid crops per dataset: Amount of valid detections whose bounding boxes are completely contained within the original picture limits.
    \item $T$ ratio: Ratio between the number of true detections -- approximated as the sum, over all pictures in the dataset, of the number of detected heads per picture, limited by the total number of annotated heads for that picture -- and the number of annotated heads in the dataset. It allows us to get an idea of the percentage of the heads of the original dataset that are correctly localised by the detector.
    \item $F$ ratio: Ratio between the number of false positives and the total number of detections. It allows us to get an idea of the percentage of false detections contained in the resulting processed dataset. 
\end{itemize}

The value we experimentally chose for $CT$ is 0.65: with this value, we can leverage more than 91\% of the AFLW dataset, and more than 97\% of  the Pointing'04 dataset -- there are no false positives for Pointing'04, but \texttildelow 40\% for AFLW. The false positives on the AFLW dataset are potentially discarded when matching the detections with the real, annotated bounding boxes, but this number of false positives could be indicative of the number of false positives that the system would find in real-world usage, and thus reducing it is still useful. The error for the AFLW dataset only decreases significantly for threshold values of 0.85 and above, and doing so has a big impact in the portion of the dataset we can use (it decreases by more than $3,000$ pictures). The values for each confidence threshold are summarized in Table \ref{table:conf_threshold}.

\begin{table*}
\centering
\vspace{1ex}
\small
\begin{tabular}{c|ccc|ccc}
\hline
\makecell{$CT$} & \makecell{\# pictures\\ AFLW} & \makecell{\textit{T} ratio\\ AFLW} & \makecell{\textit{F} ratio\\ AFLW} & \makecell{\# pictures\\ Pointing'04} & \makecell{\textit{T} ratio\\ Pointing'04} & \makecell{\textit{P} ratio\\ Pointing'04}\\
\hline
0.50 & 46686 & 0.957 & 0.420 & 46653 & 0.998 & 0.005\\
0.55 & 46174 & 0.947 & 0.414 & 46151 & 0.994 & 0.003\\
0.60 & 45520 & 0.933 & 0.408 & 45457 & 0.986 & 0.001\\
\textbf{0.65} & 44782 & 0.918 & 0.403 & 44738 & 0.974 & 0.000\\
0.70 & 43872 & 0.900 & 0.397 & 43832 & 0.959 & 0.000\\
0.75 & 42686 & 0.875 & 0.390 & 42654 & 0.934 & 0.000\\
0.80 & 41208 & 0.845 & 0.383 & 41160 & 0.895 & 0.000\\
0.85 & 39232 & 0.804 & 0.373 & 39182 & 0.831 & 0.000\\
0.90 & 36672 & 0.752 & 0.357 & 36622 & 0.723 & 0.000\\
0.95 & 32198 & 0.660 & 0.329 & 32151 & 0.511 & 0.000\\
\hline
\end{tabular}
\caption{\textbf{Confidence threshold selection.} We select $CT=0.65$, marked in bold.}
\label{table:conf_threshold}
\end{table*}

\subsection{Dataset splits} \label{ss:dataset_division}

Once we have obtained the full new dataset, we must divide it into smaller subsets before training. We make three different partitions in a stratified manner by using the class labels for each picture, obtained from its pose values (64 classes, each one containing a different combination of tilt and pan ranges). 
First, we obtain the \textit{test partition} (used for the final evaluation of trained models) as a portion containing roughly the 20\% of the entire dataset. 
Then, we obtain the \textit{validation partition} (used for tracking the training progress) as the 20\% of the remaining 80\% of the dataset. 
The rest of the dataset, after extracting test and validation partitions, will be used as the \textit{training partition}. This way, we end up with a training partition containing $57,292$ samples, a validation partition containing $14,324$ samples, and a test partition containing $17,904$ samples. These are the partitions that will be used for our experimentation and the training of our models.


\section{Neural Network architectures} \label{s:architectures}
In order to train a neural network model to solve a certain problem, we need to start from a base structure: the neural network architecture. In this work we have tested three different parameterized architectures based on the ones found in the work of Patacchiola et al. \cite{Patacchiola2017}. The authors of \cite{Patacchiola2017} present a classic convolutional architecture: a stack composed by one or more convolutional blocks (a convolutional layer followed by a pooling layer), followed by one or more fully connected layers. The activation function used (when training over the AFLW dataset) is the hyperbolic tangent. In the article the authors report an error value of 9.51\si{\degree} ($MAE$) for the pan angle over the AFLW dataset. 
In the paper, the authors use two different neural networks in order to estimate tilt and pan, respectively. This approach can be computationally demanding if we do not have a powerful system to perform the estimations, or if we require a low system-response time. 
For this reason, in our work, we aim at finding a new model capable of simultaneously estimating both angles at the same time. 
Our starting point is a modified version of the architectures proposed in their article. However, we need to accommodate a second output and to reduce the number of channels of the convolutional filters at the input from three (color) to one (grayscale).

\begin{figure}[tbp]
	\centering
	\includegraphics[width=0.75 \textwidth]{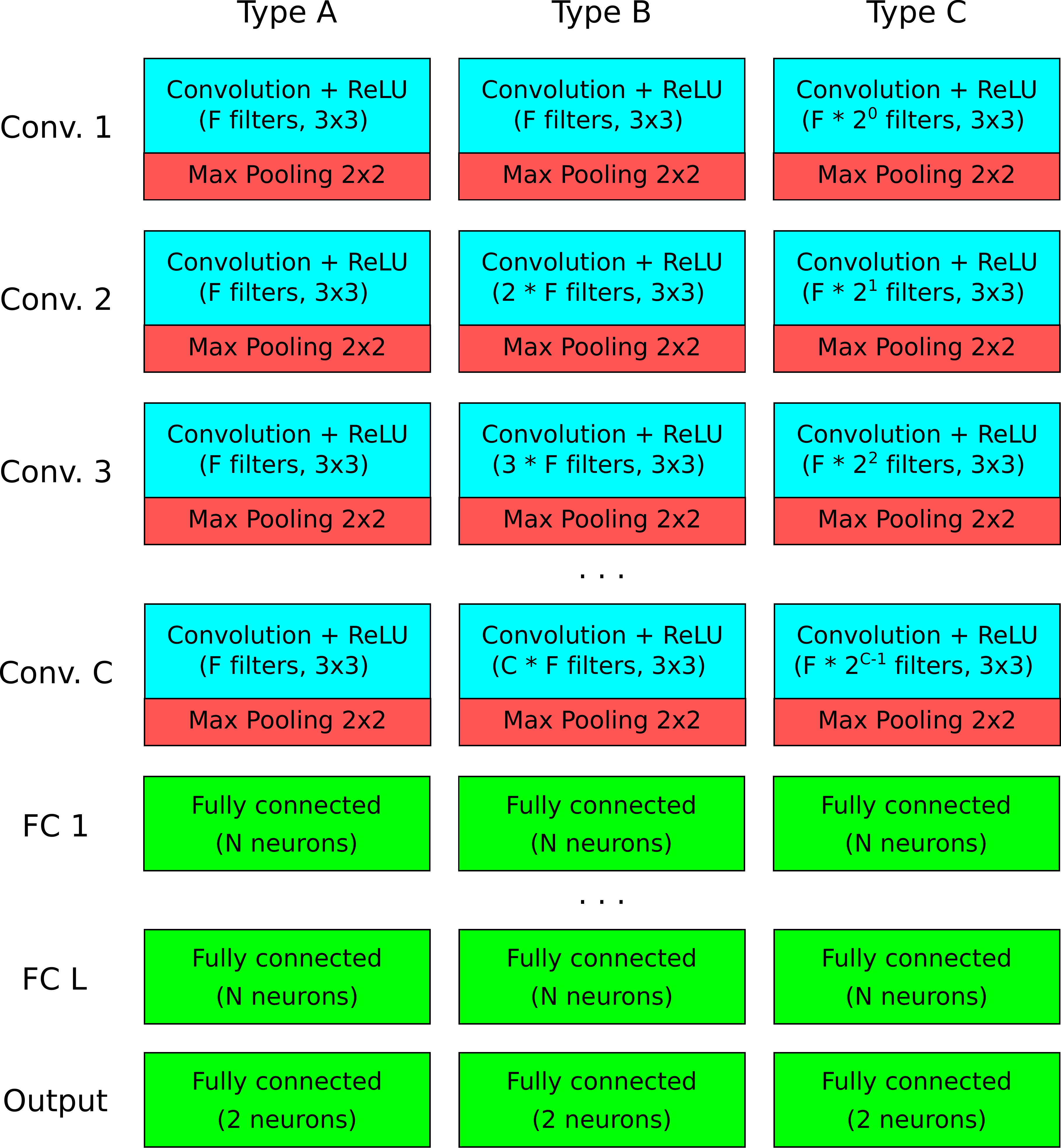}
	\caption{\textbf{Types of proposed architectures}. We propose three types of architectures to address the problem. The main difference is found in the strategy to change the number of filters in the convolutional layers. See main text for details. Acronyms: C=number of convolutional layers. Conv=convolutional layer; FC=fully-connected layer. (Best viewed in digital format.) }
	\label{fig:archs}
\end{figure}

According to the way in which the number of filters in each convolutional block varies, we experiment with three types of base architecture \rev{(see Figure~\ref{fig:archs})}:

\begin{itemize}
    \item[$\circ$] Type A: In this type of architecture the number of filters will be the same for all convolutional blocks.
    \item[$\circ$] Type B: In this type, the number of filters in each convolutional block will be progressively increased by the amount in the first block (e.g. 32 filters in the first convolutional block, 64 in the second one, 96 in the third one, etc.).
    \item[$\circ$] Type C: In this type, the number of filters in each convolutional block will be the double of the number of filters in the previous convolutional layer, starting from a base value (e.g. 32 filters in the first convolutional block, 64 in the second one, 128 in the third one, etc.). This approach is the one used by the authors in \cite{Patacchiola2017}.
\end{itemize}

The size of the input images for our family of models is $64\times 64$ pixels (see Figure~\ref{fig:flowchart}), and the size of all convolutional filters in all blocks is fixed to $3\times 3$. \rev{The last convolutional block is followed by an undetermined number of fully connected (linear) layers; the last layer of each network is a fully connected layer with two hidden units.}

\section{Experiments} \label{s:experiments}
In order to find the best possible configuration with the proposed architectures, we have carried out the experiments described in this section.

\subsection{Evaluation metrics for model comparison.} \label{ss:eval_metrics}

In order to make comparisons between different models, we will consider the following factors:

\begin{itemize}

    \item[$\circ$] Mean error: This value is computed as the mean of two values: the mean absolute error ($MAE$) for the vertical (tilt) angle, and the $MAE$ for the horizontal (pan) angle. Each of these values are obtained as the average of the error values for the corresponding metric (tilt or pan) for every pattern in the test data partition. It is expressed in degrees ($\circ$).

    \item[$\circ$] Number of floating point operations: As an indirect measure of the time a model needs to make a prediction, we can use the number of floating point operations it requires to complete a forward pass. In the case of having two models with similar mean error, we will choose the one requiring a lower number of floating point operations.
    
\end{itemize}

\subsection{Architecture search.} \label{ss:arch_search}

In order to train a neural network model, the first step is to find a suitable architecture. For each of the base architectures described in Section \ref{s:architectures}, we have tried an iterative approach, looping over multiple values for each possible parameter involved in the model definition. 
The maximum number of training epochs was set to 500, with a patience (number of epochs after which the training will stop if the validation error has not decreased) of 10 epochs. For the learning rate, we have used the default value provided by Keras for the Adam optimizer ($0.001$), reducing this value after 10 epochs if there has been no improvement (less than $0.0001$ over the validation error value). As \textit{loss} function we have used the mean squared error ($MSE$). During this initial search no data augmentation has been used. 

The tested values for each parameter are the following:
\begin{itemize}
    \item Number of convolutional blocks: from 1 to 6 (both included).
    \item Number of filters in the first convolutional layer: 32, 64, 128 and 256 filters.
    \item Number of hidden fully connected  (dense) layers: from 1 to 3 (both included).
    \item Hidden fully connected (dense) layer size: 64, 128, 256 and 512 neurons.
\end{itemize}

After completing the iterative process, we sorted the obtained results by increasing mean error; from the top results we chose the one that provided the better balance between mean error, mean prediction time and number of trainable parameters.

\subsection{Parameter search for data augmentation.}

After selecting the best architecture based on the results obtained in the previous phase, we proceed to train it over an augmented dataset: multiple transformations will be applied to the pictures in the training dataset at training time, in order to artificially increase the number of samples, reducing overfitting and, thus, improving generalization proficiency on the trained model. 
In this second phase, we have used the same iterative approach used when searching for architectures, but looping over multiple values for each possible parameter used in data augmentation. For the rest of training parameters (number of epochs, patience, learning rate, etc.) we have kept the same values of the previous phase.

The following values have been tested:
\begin{itemize}
    \item Shift range: Displacing the pixels up to 0\% (no change), 10\%, 20\% or 30\% over the length of the side of the pictures used in training (this transformation can be applied both vertically or horizontally, or even both at the same time).
    \item Brightness: Increasing or decreasing brightness up to 0\% (no change), 25\% (between 0.75 and 1.25 times the original brightness) or 50\% (between 0.5 and 1.5 times the original brightness).
    \item Zoom: Increasing or decreasing picture size up to 0\% (no change), 25\% (between 0.75 and 1.25 times the original size) or 50\% (between 0.5 and 1.5 times the original size). The final cropped picture size is the same as before applying the transformation: only the central area of the resized picture is used.
    
\end{itemize}

\section{Results}\label{s:results}

We discuss here the results obtained in the experiments described in the previous section, presenting both quantitative (on the test partition of AFLW) and qualitative results (on a diversity of videos downloaded from YouTube).

\subsection{Architecture search} \label{ss:arch_search_res}

In the first phase of the experimentation we have tried to find a good architecture, using as decision parameters both the score (the lower the error, the better) and the number of floating point operations required per estimation (the lower the number, the better). We use the data partitions described in Section \ref{ss:dataset_division}. The obtained results for each evaluated architecture are the following.

\subsubsection{Using the same number of filters in each convolutional block (type A)} \label{sss:same}

The results obtained for this type of architecture are summarized in Table \ref{table:architectures_a}. The results shown in the table correspond to different combinations of parameters used to configure the neural network (see Section \ref{ss:arch_search}), and have been obtained over the test partition described in Section \ref{ss:dataset_division} (combination of both Pointing'04 and AFLW datasets, containing $17,904$ samples).
The best result is obtained using six convolutional blocks with 256 filters per convolutional layer, followed by two fully connected   layers with 256 neurons per layer, with a mean error of 5.2\si{\degree}. 
However, when taking into consideration the amount of floating point operations needed at inference time (\ie forward pass), there is a noticeable decrease when it is used six convolutional blocks with 128 filters per convolutional layer, followed by three fully connected   layers with 512 neurons per layer (the amount of error is similar, 5.64\si{\degree}, less than 0.5\si{\degree} worse). In contrast, the number of floating point operations required decreases from roughly 1.64 billion to less than 417 million (\texttildelow $4\times$ less). Thus, this will be chosen as the best architecture of this type.

\begin{table*}[ht]
\centering
\vspace{1ex}
\small
\begin{tabular}{cccc|c|c}
\hline
\makecell{\# conv. layers} & \makecell{\# filters on\\ 1st conv. layer} & \makecell{\# dense\\ (hidden) layers} & \makecell{Dense (hidden)\\ layer size.} & \makecell{Mean\\ error (\si{\degree})} & \makecell{\# million flop\\ (rounded)}\\
\hline
6 & 256 & 2 & 256 & 5.2 & 1637\\
6 & 256 & 2 & 512 & 5.44 & 1638\\
6 & 256 & 3 & 64 & 5.5 & 1637\\
6 & 256 & 1 & 64 & 5.51 & 1637\\
6 & 256 & 2 & 128 & 5.51 & 1637\\
6 & 256 & 1 & 256 & 5.57 & 1637\\
6 & 256 & 3 & 512 & 5.61 & 1639\\
\textbf{6} & \textbf{128} & \textbf{3} & \textbf{512} & \textbf{5.64} & \textbf{417}\\
6 & 256 & 3 & 128 & 5.67 & 1637\\
6 & 128 & 1 & 512 & 5.69 & 415\\
\hline
\end{tabular}
\caption{\textbf{Architectures type A: same number of filters per convolutional block.} Ten best architectures found. The selected one is marked in bold.}
\label{table:architectures_a}
\end{table*}

\subsubsection{Increasing the number of filters per convolutional layer (type B)} \label{sss:increase}
The results obtained for this type of architecture are summarized in Table \ref{table:architectures_b}. In terms of estimation error, the best result is obtained by the following configuration: six convolutional blocks with 128 filters per convolutional layer, followed by a single fully connected   layer with 128 neurons, with a mean error of 5.34\si{\degree}. Nevertheless, considering the number of floating point operations, the architecture chosen as the best, for this type, is the second one by score, with a mean error of 5.45\si{\degree}. 
This architecture uses six convolutional blocks, with 128 filters in the first convolutional layer, followed by two fully connected   layers with 512 neurons per layer. The required number of floating point operations  decreases from nearly 1.45 billion to less than 366 million (\texttildelow$5\times$ less). Compared with the best type A architecture, this architecture achieves lower error (5.45\si{\degree} versus 5.64\si{\degree}) requiring less floating point operations (\texttildelow366 million vs \texttildelow417 million).

\begin{table*}
\centering
\vspace{1ex}
\small
\begin{tabular}{cccc|c|c}
\hline
\makecell{\# conv. layers.} & \makecell{\# filters on\\ 1st conv. layer.} & \makecell{\# dense\\ (hidden) layers.} & \makecell{Dense (hidden)\\ layer size.} & \makecell{Mean\\ error (\si{\degree}).} & \makecell{\# million flop\\ (rounded).}\\
\hline
6 & 128 & 1 & 128 & 5.34 & 1446\\
\textbf{6} & \textbf{64} & \textbf{2} & \textbf{512} & \textbf{5.45} & \textbf{366}\\
6 & 64 & 2 & 256 & 5.48 & 365\\
6 & 128 & 1 & 64 & 5.51 & 1445\\
6 & 128 & 2 & 64 & 5.52 & 1445\\
6 & 64 & 1 & 128 & 5.54 & 364\\
6 & 64 & 1 & 512 & 5.54 & 365\\
6 & 64 & 1 & 256 & 5.55 & 365\\
6 & 256 & 2 & 64 & 5.55 & 5758\\
6 & 128 & 2 & 512 & 5.56 & 1448\\
\hline
\end{tabular}
\caption{\textbf{Architectures type B: increasing the number of filters in each convolutional block by the number of filters in the first block.} Ten best architectures found. The selected one is marked in bold.}
\label{table:architectures_b}
\end{table*}
\subsubsection{Duplicating the number of filters in the previous block (type C)} \label{sss:duplicate}

The results obtained for this third type of architecture are presented in Table \ref{table:architectures_c}. Considering only the mean error, the best architecture has six convolutional blocks with 64 filters in the first convolutional block, followed by two fully connected   layers with 256 neurons per layer. This architecture achieves a mean error of 5.62\si{\degree}, requiring about 813 million floating point operations. However, other architecture gives a similar error value (5.69\si{\degree}) with just above 206 million floating point operations (\texttildelow $4\times$ less), and thus will be chosen as the best for this type of architecture. It uses six convolutional blocks, with 32 filters in the first convolutional layer, followed by a single fully connected   layer with 512 neurons. Compared with the best architecture for type B, there is an additional improvement: this architecture has a similar error (5.69\si{\degree} versus 5.45\si{\degree}), but requires far less floating point operations (\texttildelow206 million vs \texttildelow366 million).

\begin{table*}
\centering
\vspace{1ex}
\small
\begin{tabular}{cccc|c|c}
\hline
\makecell{\# conv. layers.} & \makecell{\# filters on\\ 1st conv. layer.} & \makecell{\# dense\\ (hidden) layers.} & \makecell{Dense (hidden)\\ layer size.} & \makecell{Mean\\ error (\si{\degree}).} & \makecell{\# flops.}\\
\hline
6 & 64 & 2 & 256 & 5.62 & 813\\
6 & 128 & 3 & 512 & 5.62 & 3243\\
\textbf{6} & \textbf{32} & \textbf{1} & \textbf{512} & \textbf{5.69} & \textbf{206}\\
6 & 64 & 3 & 256 & 5.69 & 814\\
6 & 64 & 2 & 128 & 5.7 & 812\\
6 & 128 & 3 & 128 & 5.7 & 3235\\
6 & 32 & 1 & 256 & 5.71 & 205\\
6 & 32 & 2 & 512 & 5.72 & 207\\
5 & 128 & 2 & 128 & 5.73 & 2482\\
6 & 32 & 1 & 128 & 5.74 & 205\\
\hline
\end{tabular}
\caption{\textbf{Architectures type C: duplicating the number of filters in each convolutional block.} Ten best architectures found. The selected one is marked in bold.}
\label{table:architectures_c}
\end{table*}

\begin{figure*}[b]
	\centering
	\includegraphics[width=\textwidth]{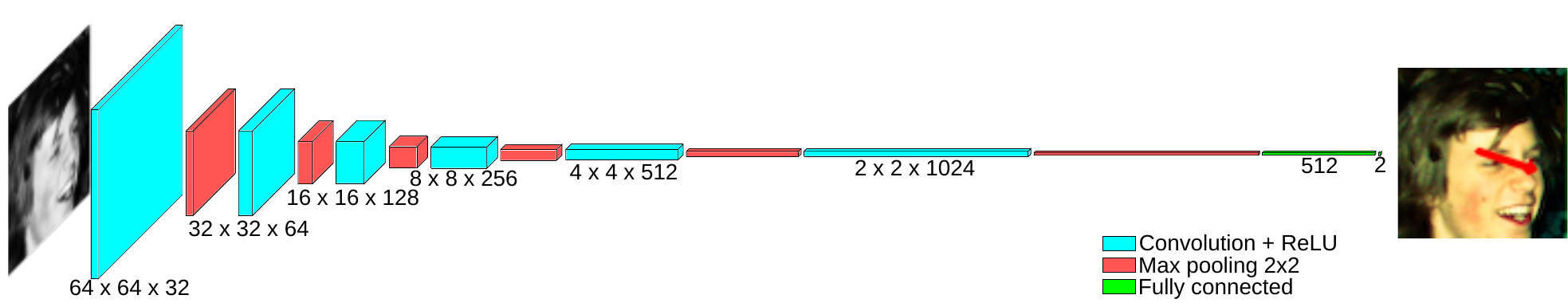}
	\caption{\textbf{RealHePoNet neural network architecture.} The input is a grayscale image of 64x64 dimensions containing a head, and the output is the estimation of both the pan and tilt angles (represented as a red arrow). (Best viewed in digital format).}
	\label{fig:RealHePoNet}
\end{figure*}

\subsubsection{Best architecture found} \label{sss:best_arch}

From the results obtained for each type of architecture, and given that the error values are quite similar (with less than 0.25\si{\degree} difference between the best and the worst) we conclude that the best search strategy is as follows: setting the number of filters for each block by duplicating the number of filters in the previous block, using a total of six convolutional blocks, starting with 32 filters on the first convolutional layer, followed by a single fully connected hidden layer containing 512 neurons (as seen in Table \ref{table:architectures_c}). The criterion used to choose such architecture is the number of flops; given that this architecture is the one requiring less operations, it should be the fastest one at inference time (\ie during pose estimation), while keeping a low error value.

\subsection{Data augmentation} \label{ss:data_augmentation}
Using the best architecture found (six convolutional blocks, starting with 32 filters on the first convolutional layer, followed by a single fully connected   layer containing 512 neurons), we aim at reducing the estimation error by training a new model over an augmented training set. As previously done, we use the data partitions described in Section \ref{ss:dataset_division}. 
In Table \ref{table:data_augmentation}, the obtained results are presented sorted by increasing mean error. The best score is achieved by allowing a max brightness variation of 50\% (\ie $0.5-1.5$) while not using any shift (\ie 0.0) or zoom (\ie 1.0). The mean error obtained with these values is as low as 4.37\si{\degree}.

\rev{We observe that the best results are obtained by simply changing the brightness of the training samples, what makes the model robust to variations in the illumination. The second best perturbation that can be applied is a change in scale (\ie zoom), what makes the model more flexible to different bounding box shapes returned by the head detector. 
Finally, perturbations based on shifting the bounding boxes a percentage of the box size (\ie left, right, top, bottom) brings only small improvements to the trained model.}

\begin{table*}
\centering
\vspace{1ex}
\small
\begin{tabular}{ccccc|c}
\hline
\makecell{Shift\\ range} & \makecell{Min\\ brightness} & \makecell{Max\\ brightness} & \makecell{Min\\ zoom} & \makecell{Max\\ zoom} & \makecell{Mean\\ error (\si{\degree})}\\
\hline
\textbf{0.0} & \textbf{0.5} & \textbf{1.5} & \textbf{1.0} & \textbf{1.0} & \textbf{4.37}\\
0.0 & 0.75 & 1.25 & 1.0 & 1.0 & 4.4\\
0.0 & 0.5 & 1.5 & 0.75 & 1.25 & 4.83\\
0.0 & 0.75 & 1.25 & 0.75 & 1.25 & 4.86\\
0.1 & 0.75 & 1.25 & 1.0 & 1.0 & 4.96\\
0.1 & 0.5 & 1.5 & 1.0 & 1.0 & 5.04\\
0.2 & 0.75 & 1.25 & 1.0 & 1.0 & 5.13\\
0.0 & 0.5 & 1.5 & 0.5 & 1.5 & 5.16\\
0.3 & 0.75 & 1.25 & 1.0 & 1.0 & 5.21\\
0.0 & 0.75 & 1.25 & 0.5 & 1.5 & 5.25\\
\hline
\end{tabular}
\caption{\textbf{Evaluation of parameters for data augmentation.} The right column shows the mean absolute error obtained by the 10 configurations that reported the lowest errors.}
\label{table:data_augmentation}
\end{table*}

\subsection{\rev{Configuration of the best model found}} \label{ss:best_model}

In summary, as reflected in the Tables of the Sections \ref{ss:arch_search_res} and \ref{ss:data_augmentation}, the best model found achieves a mean error of just 4.37\si{\degree}, \ie 3.58\si{\degree} mean tilt error and 5.17\si{\degree} mean pan error. The architecture of such model is as follows:
\begin{itemize}
    \item Type C (duplicating the number of filters in the previous block).
    \item 6 convolutional blocks.
    \item 32 filters in the first convolutional layer.
    \item 1 fully connected hidden layer.
    \item 512 neurons per fully connected hidden layer.
    \item No dropout.
\end{itemize}

\noindent During training, the following data augmentation parameters were used:
\begin{itemize}
    \item Shift range: No shift.
    \item Brightness range: Between 0.5 and 1.5 times the original brightness.
    \item Zoom range: No zoom.
\end{itemize}

\noindent Every other training parameter has been unmodified:
\begin{itemize}
    \item Batch size: 128 pictures.
    \item Maximum number of epochs: 500 epochs.
    \item Patience before training stop: 10 epochs.
    \item Learning rate: 0.001 (default value for Adam optimizer in Keras), multiplying the value by 0.1 after 10 epochs without an improvement of at least 0.0001 on the validation error.
\end{itemize}

We name this final network as \textit{RealHePoNet} (see Figure~\ref{fig:RealHePoNet}). 
\rev{In our opinion, the mean error obtained by this network is low enough for its use in many real-world applications, where images and videos are not captured in laboratory conditions.} We confirm our intuition with the results shown later in Section \ref{ss:qual_results}.

\subsection{Comparison to prior works} \label{ss:res_comparison}
\rev{We compare here the performance of our model with respect to previously published methods whose code was publicly available.}

\noindent\rev{\textbf{Comparison with Deepgaze~\cite{Patacchiola2017}.}}\\
The following comparison must be taken as merely indicative of the performance of our model with respect to the results in the original work by Patacchiola et al. \cite{Patacchiola2017}, as dataset processing and model selection strategies used in this work differ significantly in terms of approach from the ones in that work. 
For this experimental comparison, we obtained the models pre-trained by Patacchiola et al. from the public Deepgaze library repository \cite{Deepgaze}. 
In order to perform a fair comparison, in all cases, we use the same set of data, making only the changes required for each experiment.

Firstly, we compare their model against ours over the same test partition we used during our previous experiments. Note that the models by Patacchiola et al. will be tested by using the full color version of the pictures, while our model will be tested over the same ones but converted to grayscale, as required by our model. 
We have tested both approaches over AFLW dataset (the fraction of the dataset contained within the test partition, \ie $8,971$ head images). 

\begin{figure}[t]
	\centering
	\includegraphics[width=\textwidth]{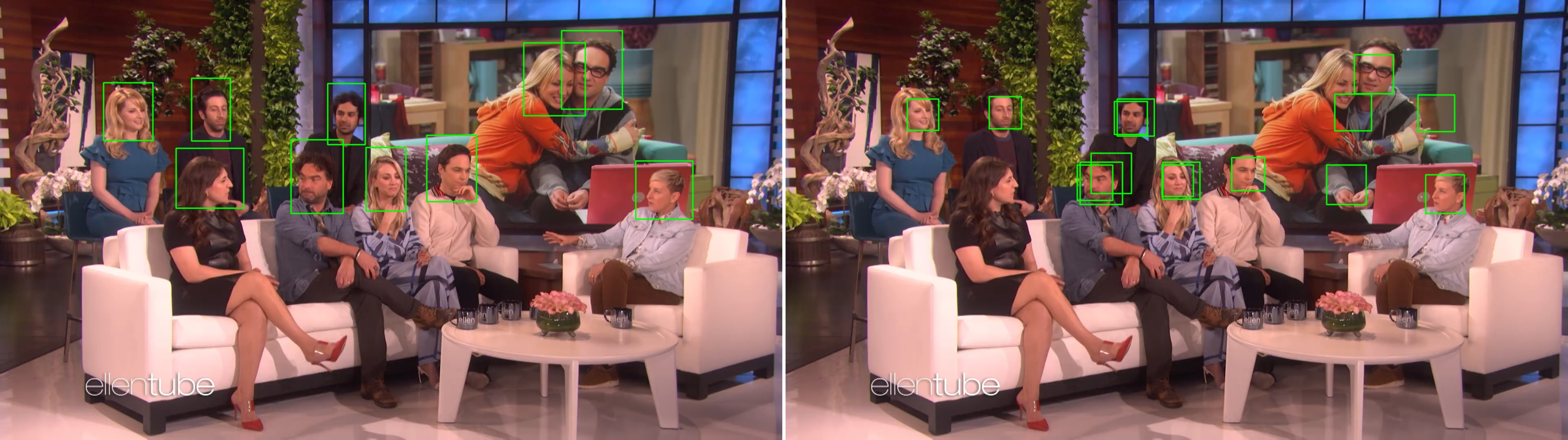}
    \caption{\textbf{Comparison between detections found with the model in the work by Marín-Jiménez et al. \cite{Marin2019} (left) and the Viola-Jones detector used by Patacchiola et al. (right)} Notice the difference in precision, making the CNN-based head detector much more suitable for this task than the Viola-Jones detector (best viewed on digital format).}
	\label{fig:detector_comparison}
\end{figure}

\begin{figure*}[b]
\centering
\begin{minipage}{0.44\textwidth}
\centering
\includegraphics[width=1 \textwidth]{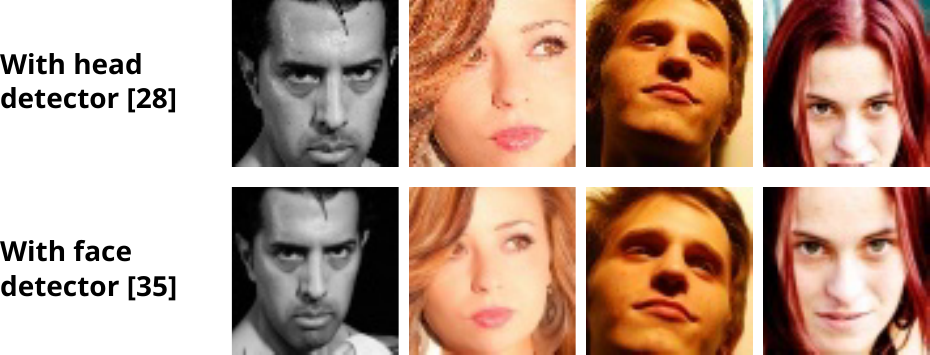}
\caption{\textbf{ Comparison between head crops obtained with our method and their counterparts obtained following~\cite{Deepgaze} (`after corrections' case).} Pictures are presented in color for better visualization; while the model by Patacchiola et al. used full color input, our model uses single-channel grayscale inputs.}
	\label{fig:comparison_img}
\end{minipage} \hfill
\begin{minipage}{0.55\textwidth}
\centering
\captionsetup{type=table} 
\footnotesize
\begin{tabular}{c|c|cc|c}
\hline
\makecell{\textit{Image}\\ \textit{conditions}}& \makecell{\textit{Model}} & \makecell{\textit{MAE}\\ tilt (\si{\degree})} & \makecell{\textit{MAE}\\ pan (\si{\degree})} & \makecell{\textit{MAE}\\ global (\si{\degree})}\\
\hline
\multirow{2}{*}{\makecell{Before\\ corrections}} & Patacchiola et al.~\cite{Patacchiola2017} & 13.33 & 31.52 & 22.43\\
& RealHePoNet (ours) & \textbf{6.88} & \textbf{9.90} & \textbf{8.39}\\
\hline
\multirow{2}{*}{\makecell{After\\ corrections}} & Patacchiola et al.~\cite{Patacchiola2017} & 9.59 & 15.40 & 12.49\\
& RealHePoNet (ours) & \textbf{7.12} & \textbf{10.77} & \textbf{8.95}\\
\hline
\end{tabular}
\caption{\textbf{Comparative results on AFLW dataset.} Mean Absolute Error (\textit{MAE}) before and after applying corrections to the person detection stage. Despite changing the detection method, our model yields more accurate estimations than its competitor.}
\label{table:comparison}
\end{minipage}
\end{figure*}

The results presented in the top row of Table~\ref{table:comparison} (`Before corrections') show a high difference in terms of mean error. Such difference might come from the pre-processing method used. Therefore, we describe below the changes applied to deal with that situation.
We can try to mimic the conditions used by Patacchiola et al. by changing the method used to downscale the cropped pictures to the final input size by using the same algorithm they used in their method -- they used `pixel area relation' instead of `bilinear interpolation'. Another difference that can be noted is the head detection model: they used their own implementation~\cite{Deepgaze} of the Viola-Jones' detector, while we are using the CNN-based head detector model from the work of Marín-Jiménez et al.~\cite{Marin2019}. 
\rev{To visually identify the possible main differences in this regard,} a comparison between the output of both detectors over the same input picture can be seen in Figure \ref{fig:detector_comparison}. \rev{We observe that our detector (left) tends to enclose the whole head of the person within the detection bounding box, using a variable aspect ratio. In contrast, the Deepgaze's  detector~\cite{Deepgaze} (right) focuses on just the face region of the head, using a fixed aspect ratio (\ie squared bounding box). In summary, our detector includes contextual information that has not been seen by Patacchiola's model during training and, therefore, it probably does not how to handle it at inference time.}

We have processed the AFLW dataset making the changes needed to adapt the inputs to the models. 
As a result, the amount of false detections have been reduced by requiring a minimum IoU value of 0.25 before declaring a match between a detected head and an annotated head (see Subsection \ref{sss:aflw_dataset}). 
Trying to rule out any possible advantage caused by testing over training data, we extracted only the heads with a matching picture in the original test partition used during our experimentation: this left us with a smaller set of $3,988$ pictures. We tested the models over this refined dataset (the one by Patacchiola et al. using color pictures and ours using grayscale pictures). 
Figure \ref{fig:comparison_img} shows a comparison between multiple pictures processed with our original method versus their counterparts obtained after applying the suggested corrections. The results of these experiments, both \textit{before} and \textit{after} applying these corrections, are summarized in Table \ref{table:comparison}.

From the comparison in Table \ref{table:comparison}, we observe that the performance of our model is competitive, compared to the CNN architectures used as reference. The differences between the performance obtained by the models by Patacchiola et al. and the performance claimed in their work \cite{Patacchiola2017} may be due to differences in the input image processing, and it should not be taken as indicative of the actual performance of their models. However, these results highlight the robustness of our model when tested outside its original training conditions (using a different head detection method).
\rev{Recall that the Patacchiola's estimator is not a single model but a set of models, one per angle. In contrast, RealHePoNet returns both angles in a single forward pass, using as input a grayscale image, what reduces the computational burden. }
The code developed for this comparison is available in our GitHub project~\cite{headpose_final}.

\noindent\rev{\textbf{Comparison with Hopenet~\cite{ruiz2018hopenet}.}}\\
\rev{
 The model proposed by the authors of Hopenet is built around the ResNet50 architecture \cite{ResNet}, a residual network composed of 50 layers, typically used for image recognition tasks. At the output of this base architecture, the authors have added three different branches, corresponding to the three different predicted angles. Each one is composed of a fully connected layer followed by a softmax output layer that tries to assign the pose to a certain angle bin. The output of these three branches can be processed individually in order to obtain fine-grained predictions.
 For our testing procedure, we have used one of the models provided by the authors in the Hopenet repository \cite{hopenet_repo} (300W-LP, alpha 1, robust to image quality); this model has been chosen as it should be the most suitable to be applied over real-world pictures as the ones appearing in the AFLW dataset. The input images correspond to the portion of the AFLW dataset used to test our model; they have been obtained using the head detector in the work by Mar\'in-Jim\'enez et al. \cite{Marin2019}, but they have been resized to $224\times224$; also, as the ResNet50 model uses color pictures as input, the pictures have not been converted to grayscale.
}

\rev{In order to evaluate the robustness of our model with respect to the face detector used to define the input, we report here the results obtained using as detector the one recommended by Hopenet: Dockerface~\cite{ruiz2017dockerface}. This detector returns a set of $3,778$ valid heads from the test partition of AFLW, using the recommended confidence detection threshold of $0.85$.
With these same conditions for both pose estimators, the average error obtained for Hopenet is 10.65º (9.42º for tilt and 11.8º for pan) and 12.10º for RealHePoNet (8.77º for tilt and 15.43º for pan). Note that the difference in precision between both methods is lower than 2º. In contrast, our model is much simpler (less parameters) and faster.
In terms of computational time, as a reference, Hopenet requires 10.8 milliseconds to produce an estimation, what is slower than our model using the same hardware (see Sec.~\ref{subsec:timing}).
}

\rev{In contrast, allowing our model to use the same head detector used during training, the error decreases significantly, from an average of 12.10º to 8.39º (\ie 6.96º for tilt and 9.83 for pan).
This means that, although our model is somehow robust to changes in the input format (\eg different margin around the face, face shift, etc.), the best performance is obtained in collaboration with its corresponding head detector.
In summary, our model improves the previous result presented for Hopenet (10.65º) reducing the average error around 2º, by using smaller grayscale inputs (\ie $64\times64\times1$ vs $224\times224\times3$).
}

%

\subsection{Qualitative results} \label{ss:qual_results}
\begin{figure}[tbp]
	\centering
	\includegraphics[width=1.0 \textwidth]{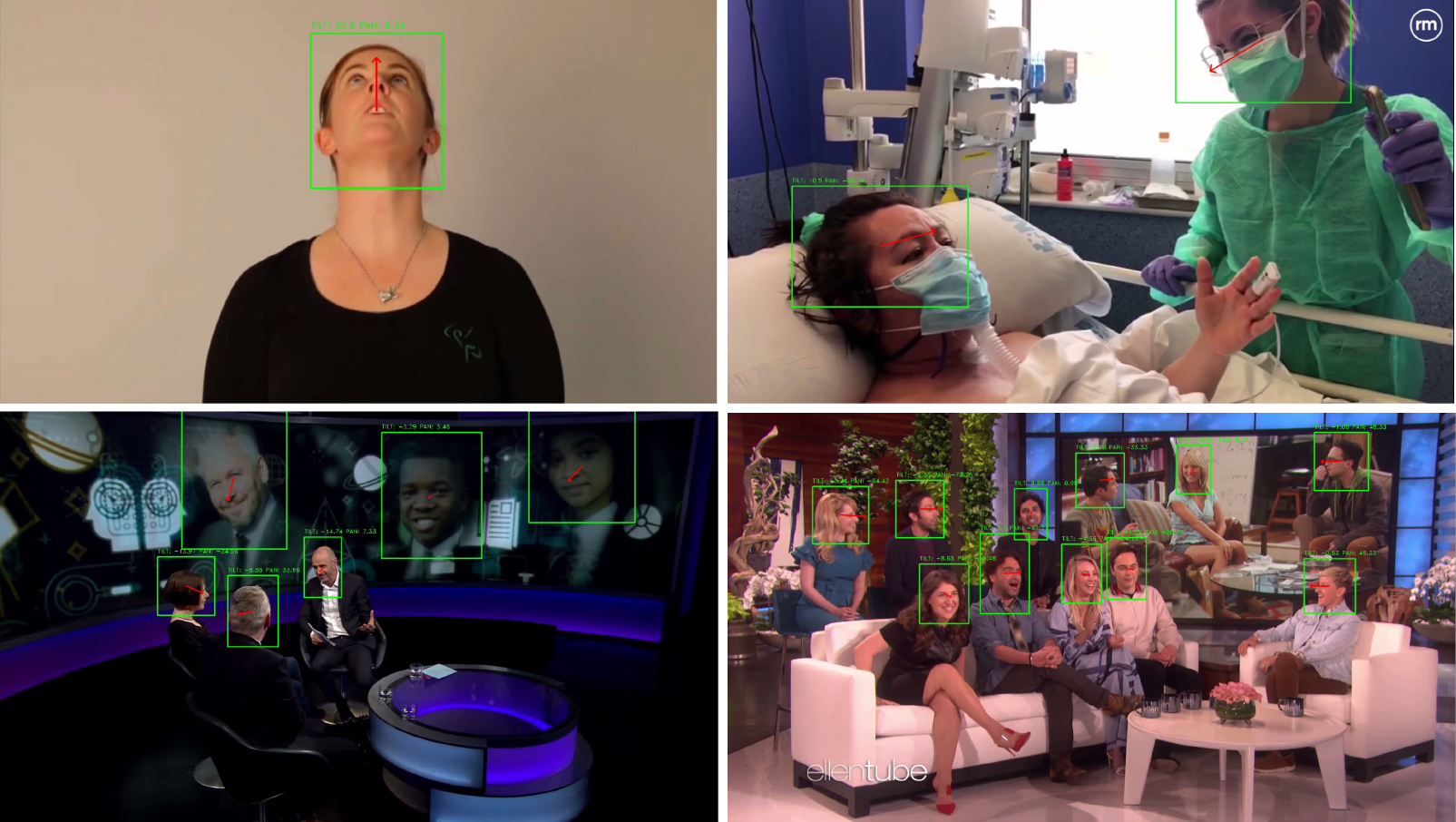}
	\caption{\textbf{Qualitative results}. Results included in the demo video generated from different YouTube videos. Note the diversity of situations where our model works without any additional fine-tuning.  (Best viewed in digital format.) }
	\label{fig:youtubesamples}
\end{figure}

\begin{figure*}[tbp]
\centering
\includegraphics[width=0.48 \textwidth]{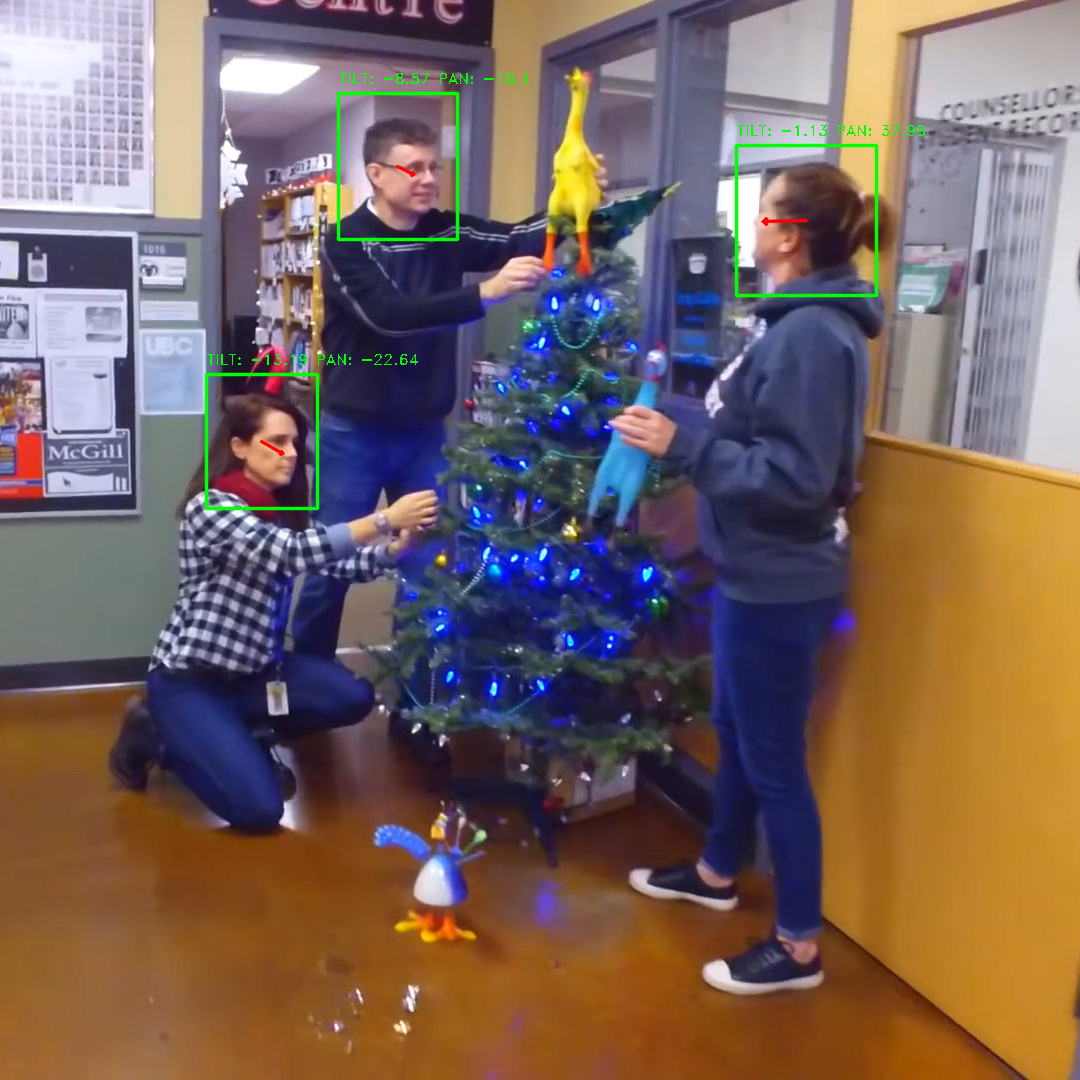}
\includegraphics[width=0.48 \textwidth]{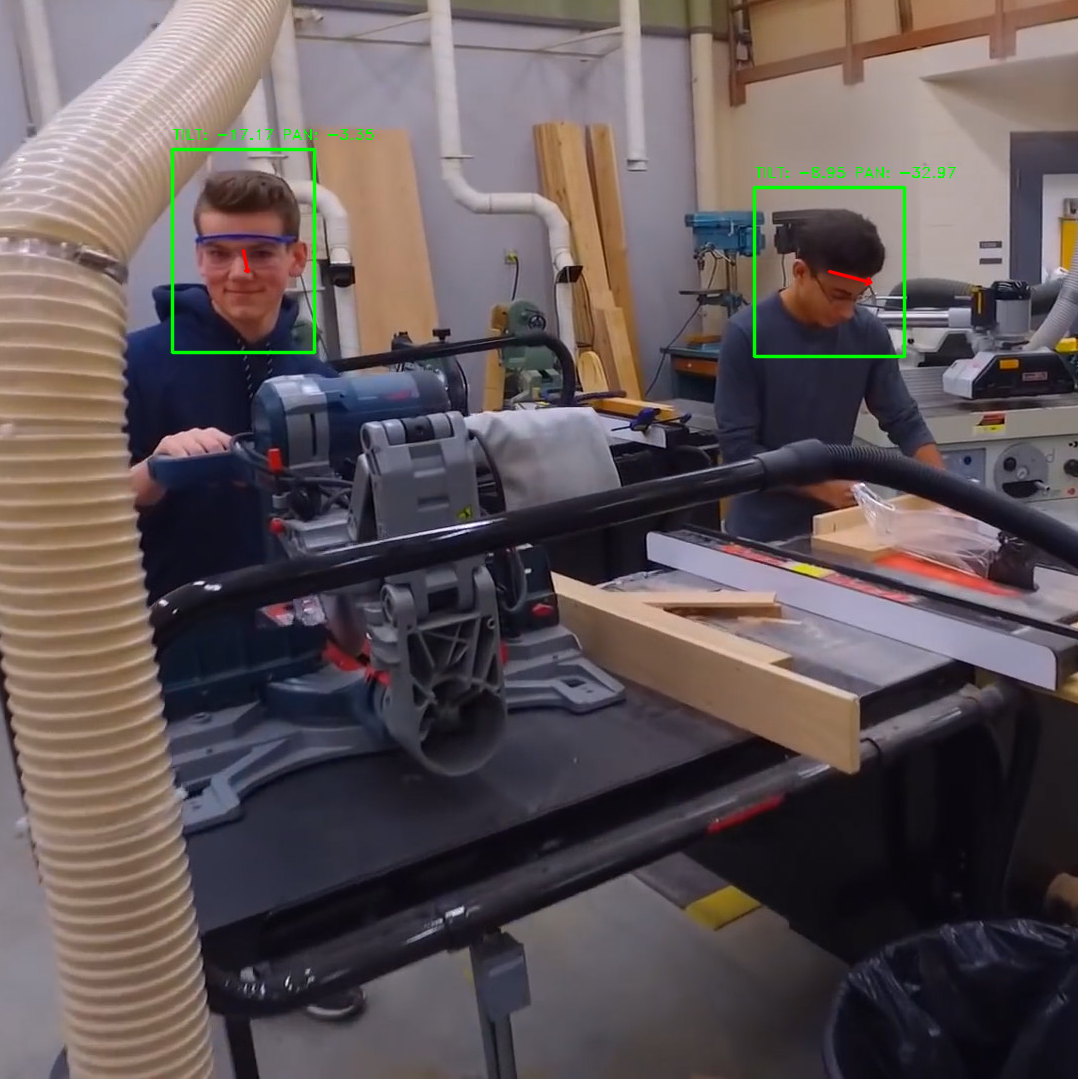}
\includegraphics[width=0.48 \textwidth]{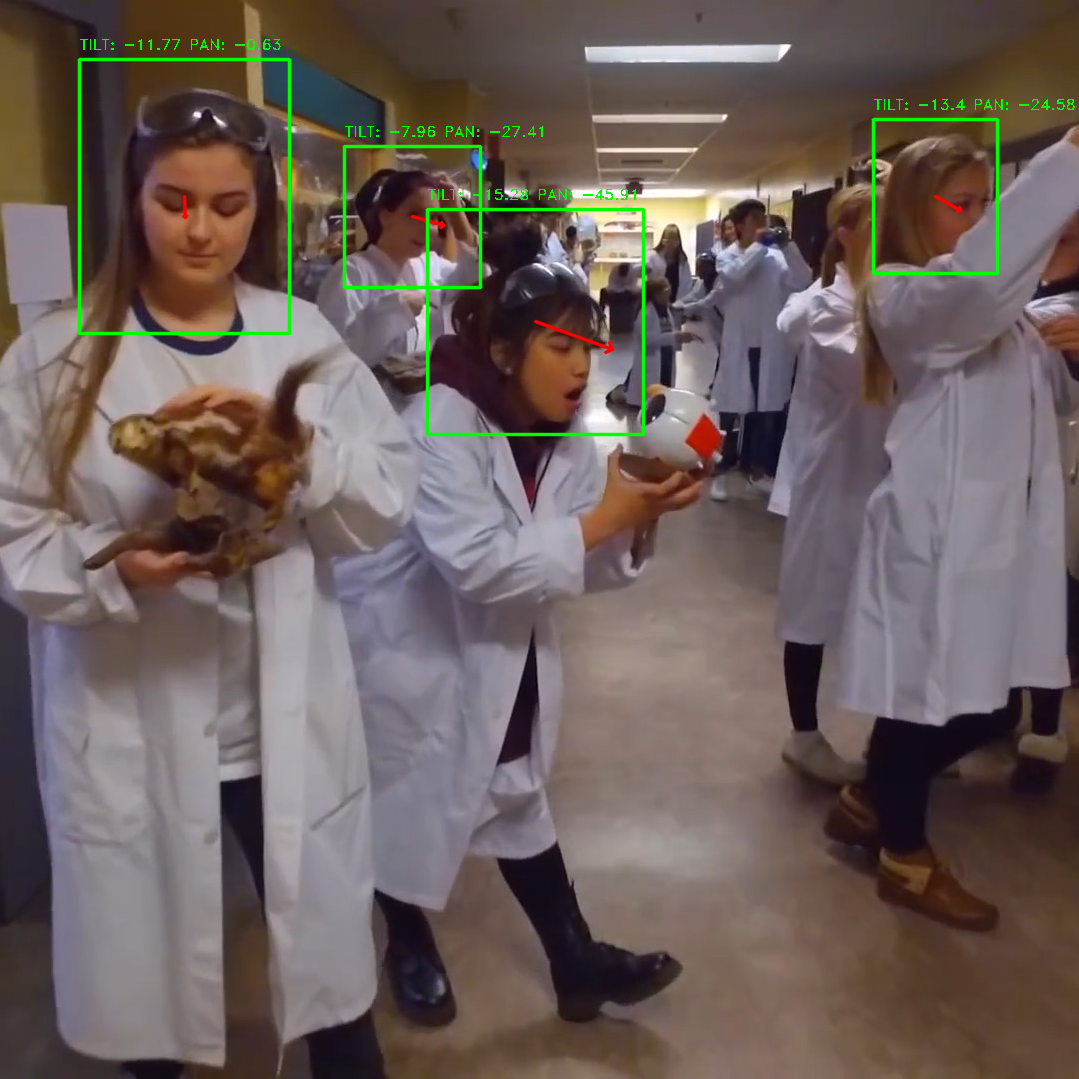}
\includegraphics[width=0.48 \textwidth]{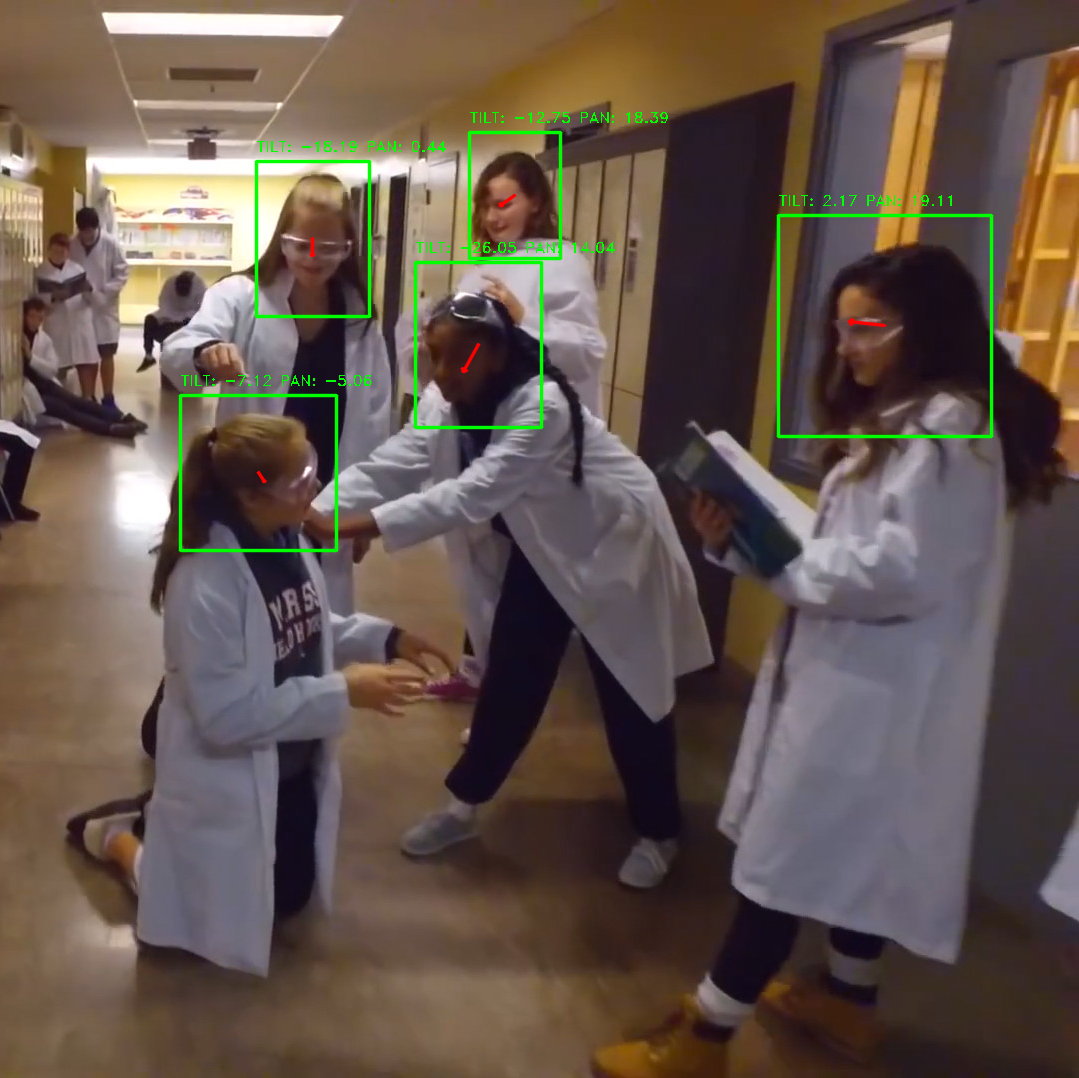}
\caption{\textbf{Predictions obtained with the trained model.} Images obtained from YouTube video \cite{Mannequin}. Despite the difficulty of this footage, most estimations look correct. In the bottom right image, we show a case (bottom left girl) where the estimation is not correct. (Best viewed in digital format).}
\label{fig:qualitative_results}
\end{figure*}

\rev{In addition to the benchmark results presented in the previous section, we present here qualitative results. That is, we are interested in visually evaluating the performance of our model on challenging images coming from totally different data domains where the model was trained. This will show the capacity of generalization obtained by the model during training and will exhibit its behaviour on real-world images.}
Pictures in Figures \ref{fig:youtubesamples} and \ref{fig:qualitative_results} are the result of directly testing RealHePoNet over a set of frames from YouTube videos.
In particular, the results in Figure \ref{fig:qualitative_results} come from the YouTube video entitled \textit{High School Mannequin Challenge 1500 Students - Maple Ridge Secondary School} \cite{Mannequin}, where the camera is moving along different corridors and rooms with people of diverse ages.
We have also prepared a demo video \cite{Demo} with additional qualitative results showing the performance of our model. Such demo video compiles processed footage from the previous test video, in addition to other four test videos \cite{BigBang, SocialMobility, Coronavirus, Neckwarmup}. Some frames of the demo video are shown in Figures~\ref{fig:teaser} and \ref{fig:youtubesamples}. 
We can observe the good behaviour of RealHePoNet in many different scenarios, with diverse number of people (up to 12 people detected simultaneously in the same scene) and scale, with images even affected by motion blur. 
These results suggest that the model has a high capacity of generalization, working on images that come from different domains never seen during training.

Note that the results shown in the demo video have not been post-processed, \ie they directly correspond to the per-frame raw output of the models (detector and pose estimator) without any temporal smoothing or filtering.

\subsection{\rev{Computational performance at inference time}}~\label{subsec:timing}
Over our combined dataset, the mean inference time measured for the trained model was \texttildelow0.26 milliseconds per head. This value was obtained using a system with a GTX 1060 GPU (1280 CUDA cores @ 1847 MHz, 6 GB VRAM). 
In order to provide a more detailed information on the computational performance of the complete system, we have tested the model performance over one full video used in the demo~\cite{SocialMobility}, re-scaling it to multiple video resolutions. The full video contains $9,343$ frames. The obtained results are summarized in Table \ref{table:performance}.

\begin{table*}[htb]
\centering
\vspace{1ex}
\small
\begin{tabular}{c|cccc}
\hline
\makecell{\textit{Resolution}} & \makecell{\textit{Mean} \#\\ \textit{of heads}} & \makecell{\textit{Mean head}\\ \textit{detection time (ms)}} & \makecell{\textit{Mean pose}\\ \textit{estimation time (ms)}} & \makecell{\textit{Avg.} \\\textit{FPS}}\\
\hline
$854 \times 480$ & 1.90 & 57.56 & 5.84 & 15.18\\
$1280 \times 720$ & 1.92 & 59.23 & 6.07 & 14.12\\
$1920 \times 1080$ & 1.93 & 59.69 & 6.15 & 12.37\\
\hline
\end{tabular}
\caption{\rev{\textbf{Computational performance of RealHePoNet for different image resolutions.} The results (per image) correspond to the average number of heads, the head detection and pose estimation time (in milliseconds) and the average frames per second (FPS) achieved during the processing of the test videos.}}
\label{table:performance}
\end{table*}

The differences in detection time are negligible, and they are most likely caused by the image resizing step needed in order to provide an input to the head detector. On the other hand, the detector is able to detect more heads over an input image re-scaled from a larger picture, thus explaining the small increase in the mean number of heads per frame and the increase in the mean pose estimation time per frame. While performance differences in pose estimation time could be explained by the computational overhead introduced by other operations required prior to the pose estimation stage when using it over non-preprocessed pictures. In any case, \texttildelow166 pose estimations can be performed per second (at a GTX 1060 GPU).
Note that the detection stage is $10\times$ slower than the head pose estimation stage (\ie the goal of this paper). Therefore, we hypothesize that obtaining a real-time head pose estimator is easily plausible by just optimizing the head detection method.



\section{Conclusion} \label{s:conclusion}

In this work, we have addressed the human head pose estimation problem with ConvNets, obtaining practical results on real-world images, in contrast to laboratory condition images.
We have carried out a methodological evaluation on a family of ConvNet architectures with the goal of finding a model showing a good trade-off between accuracy and computational burden. 

The resulting model, RealHePoNet, yields a low error value while trying to keep the number of floating point operations as low as possible, making it usable even on modest hardware. 
Part of this success is due to the combination of two datasets for training: one obtained in controlled conditions (Pointing'04) and other obtained from real-world images (AFLW). 

Both quantitative and qualitative results suggest that our model can be used in a wide variety of situations without the need of fine-tuning.
Furthermore, our iterative approach for architecture parameter search can be used to obtain a good enough model in a relatively short time, provided enough computational power.

\rev{In summary, we contribute, through a public software release, a single-stage ConvNet for head pose estimation that requires as input just a $64\times64 $ grayscale image containing a human head and returns both pan and tilt angles. Thanks to its modest amount of parameters (compared to existing methods) and, therefore, of operations, RealHePoNet is able to perform \texttildelow166 pose estimations per second (tested on a GTX 1060). 
}

\rev{
The known limitation of the system, as a whole, is the current speed of the head detector, \ie $10\times$ slower than the pose estimator. Therefore,
as future line of research, it might be of interest to couple our pose estimator with a faster and lighter head detector, what was out of the scope of this current work. Secondly, it might be worthy evaluating the computational performance of the model on mobile devices, where energy consumption, along with limited computation resources, is an issue.
And, finally, as the current model only estimates the two main components of the pose (pan and tilt), as both are enough for most real-world applications (\eg focus of attention), adding the estimation of the roll angle might be the objective of future developments.
}



%
%

\section*{Acknowledgements}

This work has been partially funded by the Spanish projects TIN2019-75279-P and RED2018-102511-T.
We gratefully acknowledge the support of NVIDIA Corporation with the donation of the Titan X Pascal GPU used for this research.

\section*{Conflicts of Interest}
The authors declare that they have no conflict of interest.

\section*{Code availability}
Code is publicly available at: \url{https://github.com/rafabs97/headpose_final/}

\section*{\rev{Notation}}
\rev{\begin{itemize}
    \item [AFLW] Annotated Facial Landmarks in the Wild.
    \item[CNN] Convolutional Neural Network.
    \item [Conv] Convolution.
    \item [ConvNet] Convolutional Neural Network.
    \item [CT] Confidence Threshold.
    \item [FC] Fully connected.
    \item [flops] Floating point operations per second.
    \item [FPS] Frames per second.
    \item [HPE] Head pose estimation.
    \item [IoU] Intersection over Union.
    \item [MAE] Mean Absolute Error.
    \item [MSE] Mean Squared Error.
    \item [SSD] Single Shot Detector.
\end{itemize}
}

\bibliographystyle{spbasic}      
\bibliography{bibliografia}   

\end{document}